\documentclass[graybox]{svmult}

\usepackage{type1cm}        
\usepackage{makeidx}         
\usepackage{graphicx}  
\usepackage{gb4e}
\usepackage{booktabs}
\usepackage{multirow}
\usepackage{dirtytalk}
\usepackage{multicol}        
\usepackage[bottom]{footmisc}
\usepackage{times}
\usepackage{xcolor}
\usepackage{float}
\usepackage{amsmath}
\usepackage{amssymb}
\usepackage{nicefrac}
\usepackage{cleveref}
\usepackage[caption=false]{subfig}

\noautomath

\newcommand{\ee}{\mathcal{E}}

\definecolor{blu}{HTML}{005082}

\makeindex             

\newcommand{\blank}{$\rule{0.6cm}{0.15mm}$}

\begin{document}

\title*{Finding Fuzziness in Neural Network Models of Language Processing}
\author{Kanishka Misra and Julia Taylor Rayz}
\institute{Kanishka Misra and Julia Taylor Rayz \at Purdue University, West Lafayette, IN 47906, USA\\\email{kmisra@purdue.edu, jtaylor1@purdue.edu}}
%
%
\maketitle

\abstract*{Humans often communicate by using imprecise language, suggesting that fuzzy concepts with unclear boundaries are prevalent in language use. In this paper, we test the extent to which models trained to capture the distributional statistics of language show correspondence to fuzzy-membership patterns.
Using the task of natural language inference, we test a recent state of the art model on the classical case of temperature, by examining its mapping of temperature data to fuzzy-perceptions such as \textit{cool}, \textit{hot}, etc. We find the model to show patterns that are similar to classical fuzzy-set theoretic formulations of linguistic hedges, albeit with a substantial amount of noise, suggesting that models trained solely on language show promise in encoding fuzziness.}

\abstract{Humans often communicate by using imprecise language, suggesting that fuzzy concepts with unclear boundaries are prevalent in language use. In this paper, we test the extent to which models trained to capture the distributional statistics of language show correspondence to fuzzy-membership patterns.
Using the task of natural language inference, we test a recent state of the art model on the classical case of temperature, by examining its mapping of temperature data to fuzzy-perceptions such as \textit{cool}, \textit{hot}, etc. We find the model to show patterns that are similar to classical fuzzy-set theoretic formulations of linguistic hedges, albeit with a substantial amount of noise, suggesting that models trained solely on language show promise in encoding fuzziness.}

\section{Introduction}
Fuzziness of variables such as \textsc{hot}, \textsc{tall}, etc. is often reflected in language use.
Humans tend to use vague constructions such as \textit{``Joe is quite tall''} or \textit{``it's very cold outside''} in their everyday language and are still able to successfully communicate their intent.
Fuzzy sets \cite{zadeh1965fuzzy} provide us with a calculus for handling such imprecise expressions, and have been used in numerous engineering applications, but have yet to be fully adopted by the Natural Language Processing (NLP) community \cite{carvalho2012critical}. Computing with Words (CWW) \cite{zadeh1999fuzzy} initiatives followed a similar trend: they are described by engineers and computer scientists and used in some applications, but are not trending in publications of natural language processing (NLP) fields. 

A prevalent hypothesis in the computational linguistics and natural language processing communities is that the distributional statistics of language use can guide the process of acquiring word meaning --- that \textit{``You shall know the meaning of a word by the company it keeps''} \cite{firth1957synopsis}.
This distributional hypothesis has been the basis of a number of prominent language representation architectures, ranging from early neural network-based vector space models such as \texttt{word2vec} \cite{mikolov2013distributed} to more recent models that fall under the category of \textit{``pre-trained language models,''} such as BERT \cite{devlin-etal-2019-bert}.
A key factor in models that follow the distributional hypothesis is that their knowledge is acquired solely from linguistic input, and reflects the extent to which the statistics present in the language can inform the model about the world.
Indeed, recent studies have shown models such as BERT to show great feasibility in capturing factual \cite{petroni2019language} and categorical knowledge \cite{weir2020probing} by investigating their outputs on word prediction prompts such as \textit{``A robin is a \blank{} .''}
While these analyses shed light on how well prediction in such models reflects what is generally known to be true in the world (in a bivalent setting of strict truth and false), much is unknown about the manifestation of vagueness within the semantic knowledge that they capture.

We argue in this paper that models that are built to interpret and act upon language should also be able to account for vagueness in categories and predicates, which have borderline cases where the exact truth value is unclear.
This is especially important in the case of systems involving rational human interaction, where, for example, a computer can communicate to the user how \textit{icy} a road is on a particular wintry night by approximating it from precise data \cite{van2012not}, and for this to be made possible, our models of language processing must encode aspects vagueness or fuzziness.
To test this, we investigate a state of the art NLP model (RoBERTa$_{\textit{large}}$ \cite{liu2019roberta}) in its ability to encode fuzziness of language, specifically in this case, language about a classical fuzzy concept: \textsc{temperature}.
We formulate the probing experiment as a Natural Language Inference (NLI) task where the model is supplied with precise descriptions about temperature data and the model is evaluated based on the extent to which that data maps to opportunistically selected fuzzy categories such as \textit{freezing, warm, etc.}


\section{The concept of vagueness and the vagueness of concepts}
\label{sec:vagueness}
We define vagueness to be the phenomenon underlying the uncertainty in demarcating concepts, i.e., the lack of clear-cut boundaries separating membership to one concept with the other.

\subsection{Psychological significance}
Experimentally, vagueness has been of interest in the psychological study of concepts.
The seminal work of Rosch \cite{rosch1975cognitive} showed consistently that people perceived category membership as a graded phenomenon, as opposed to being a matter of crisp boundaries. 
Labov \cite{labov1973boundaries} experimented on the boundaries of commonplace concepts such as \textsc{cup}, \textsc{mug}, and \textsc{bowl}, by presenting human subjects with exemplar drawing of the objects.
He found subjects to be consistent in attributing names to object drawings with typical visual characteristics, while showing large inconsistencies between each other while naming drawings where the object was visually intermediate between categories (for instance being too tall to be a member of \textsc{bowl} but too wide to be a \textsc{cup}), indicating the close relationship of lack of consensus with vagueness.
Additionally, the work of McCloskey and Glucksberg \cite{mccloskey1978natural} involved experiments where human subjects had to explicitly rate whether a particular category (e.g. \textsc{fruit}) applied to multiple items in consideration (e.g. \textsc{apple, banana}). 
The authors found that when the same group of people were asked to perform the experiment again, after a few weeks, the subjects tended to be highly inconsistent with their original judgements for borderline cases (e.g. \textsc{tomato}), suggesting that vagueness also involves individual uncertainty.

\subsection{Handling vagueness using fuzzy interpretations of linguistic hedges}
\begin{figure}
\sidecaption[t]
    \centering
    \includegraphics[width = 0.5\textwidth]{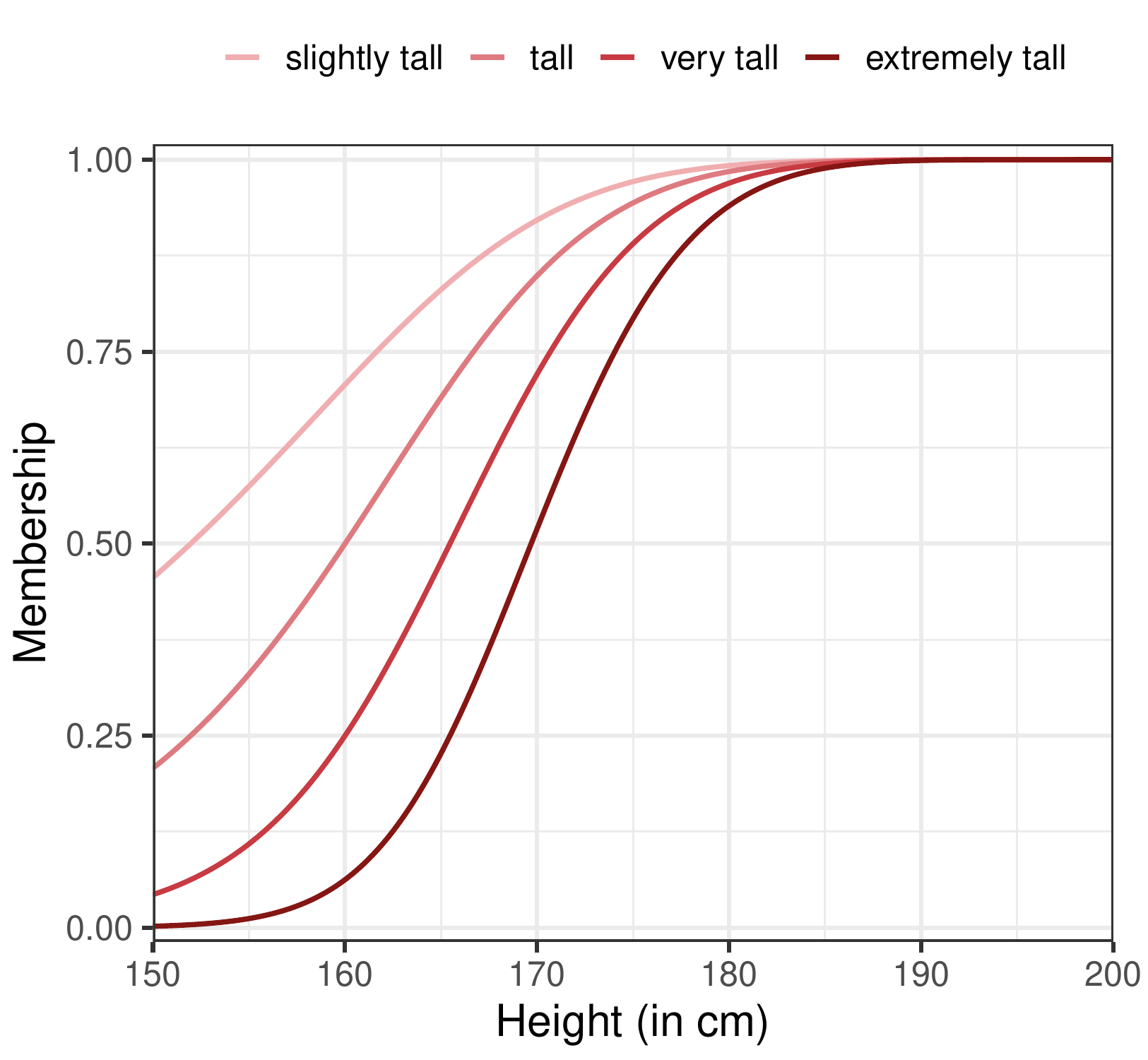}
    \caption{Possible membership transformations for linguistic hedges of \emph{tall} (with membership function $\mu(h)$, generated using a general bell curve): (1) \textit{slightly tall} -- $\mu^{1/2}(h)$; (2) \textit{very tall} -- $\mu^2(h)$; and (3) \emph{extremely tall} -- $\mu^4(h)$, where $h$ is height. Note that this is solely for illustrative purposes and is not an exhaustive list of hedges of \textit{tall}.}
    \label{fig:example}
\end{figure}

Linguistic hedges \cite{lakoff1975hedges}, such as \textit{sort of, technically, etc.,} are central to vagueness in language.
Zadeh's seminal work \cite{zadeh1965fuzzy} proposes a calculus to handle such fuzziness in scalar adjectives such as \textit{tall}, \textit{old}, etc.
Subsequent work \cite{zadeh1972fuzzy} proposed to explicitly treat hedges, such as \textit{very, more-or-less, slightly} as operations on pre-defined fuzzy sets.
Usually, the definition of the membership functions and fuzzy sets is left to the experts, as it is one of the most difficult tasks in a successful fuzzy expert systems \cite{martine2002linguistic}.
For instance, Zadeh proposed that for a given fuzzy set $F = \{(x, \mu_F(x))\}$, hedges such as \textit{very, extremely, etc.} can be modeled using ``concentration'' functions that pull the curve of memberships for $F$ inward, such as $\textit{very }F = \mu_F^2(x)$.
Meanwhile, hedges such as \textit{slightly, almost} were modeled using ``dilation'' functions that spread out the values of the membership function, such as $\mu_F^{1/2}(x)$.
Zadeh also proposed more complex interpretation of hedges, by employing the use of ``intensifiers.''
An example of applying such recommendations to the concept of tall is shown in \cref{fig:example}. 
Whether or not Zadeh's proposal for fuzzy-interpretations of linguistic hedges is correct, they provide a useful method for handling vague predicates in semantics and serve as inspiration for some of our analyses.

While fuzzy sets in general, and linguistic hedges---modeled as operators---are defined by experts to model data, we wonder whether very large amount of data can model such linguistics hedges. 
Given a membership function, and some point $x$, one can determine whether this point $x$ is a member of a fuzzy set, and to what degree.
What we are interested in is an answer to this question: given a point $x$, and approximate membership value of some fuzzy set, is it possible to reconstruct a fuzzy membership function for $x \in X$ where $|X|$ is very large.

In other words, we can assume that if somebody says that it is 90$^\circ$F outside, we find a membership degree of \textit{hot, very hot, slightly hot, etc.} using an expert-determined fuzzy membership function. 
What is less clear is whether an oracle (let us assume that we have such a thing) would be able to map 90$^\circ$F to some relative position of \textit{hot}, \textit{very-hot}, etc. without \textit{knowing} the actual equations for the membership functions.

More specifically, we wonder if the notion of fuzzy sets and hedges---modeled as proposed by Zadeh---can be reconstructed from very large data. If such reconstruction can be done, we should see graphs that are qualitatively similar to that of \cref{fig:example}.
However, it should be noted that the linguistic hedges are indeed linguistic in nature, and thus, they can allow synonym substitutions. 
In particular, somebody very very tall may be called a giant, and our data may show correlation with that word choice, but not with ``very very tall."
While this raises the complexity of analysis, we note that it is essential in order to facilitate conversational interactions between computers and humans, but leave the fine-grained details for future work.

\section{Pre-trained Language Models}
Recent advances in the field of NLP have lead to the rise of a class of highly parameterized neural network-based models called pre-trained language models (PLMs).
These models are trained on vast amounts of text using the language modeling objective --- estimating probabilities of missing words in context.\footnote{we assume a general form of the language modeling objective which also includes its bidirectional sense}
A common architecture that governs the computation in these models is the transformer architecture \cite{vaswani2017attention}.
The attention module in the transformer architecture allows PLMs to `remember' all words in a given sentence context, while predicting the missing word, thereby making its internal representations more finely attuned to the entire context.
Though originally proposed for ranking sentences (by assigning them probabilities), the language modeling objective guides PLMs into learning general-purpose language representations which can be fine-tuned to any supervised learning task end-to-end, i.e., all representations get updated in order to optimize for a given task(s). 
Their expressiveness and adaptability has enabled PLMs to achieve state of the art results on several high-level NLP tasks, such as Question Answering, Natural Language Inference, Reading Comprehension, etc. 
PLMs come in two main variants: (1) Incremental PLMs, which are trained to predict words when conditioned on a left context, e.g., \textit{``I went to the library to \blank{}''}; and (2) Masked PLMs, which have bidirectional context while predicting tokens, e.g., \textit{``I went to the \blank{} to read.''}
Examples of Incremental PLMs include GPT2 \cite{radford2019language} while those of Masked PLMs include BERT \cite{devlin-etal-2019-bert} and RoBERTa \cite{liu2019roberta}.

\section{Vagueness in Natural Language Inference Models}
In this section, we present our behavioral experiment targeting the extent to which NLP models, trained on very large data, are sensitive to fuzziness/vagueness of concepts.
We specifically consider the classical fuzzy concept of \textsc{temperature}, where, for instance, the transition from \textit{cool} to \textit{warm} is gradual, and there is no clear-cut separation between when someone perceives their surrounding to fall under either of the two categories.
To probe for how models encode the fuzziness of \textsc{temperature}, we rely on the task of natural language inference (NLI) \cite{bowman2015large}. 

NLI, also referred to as ``recognizing textual entailment'' (RTE), is the task of predicting the logical entailment relation between pairs of sentences.
Specifically, a model trained to do this task is supplied with two sentences, a premise and a hypothesis, and it has to predict whether the premise entails the hypothesis.
For instance, the sentence \textit{``a dog is running towards the man''} logically entails \textit{``an animal is moving towards a human''}, while \textit{``the boy is mowing the lawn''} contradicts \textit{``the boy is sleeping''}.
Note that it is assumed that the premise and the hypothesis must refer to the same event in order for this task to be carried out.
NLI features prominently in many language processing benchmarks \cite{wang2018glue, wang2019superglue}, highlighting its importance in progress towards achieving true natural language understanding, and by extension, an understanding of the world.

We argue that the task of NLI is suitable in our endeavor of exploring fuzziness in models due to the amount of flexibility it offers. 
Models trained to perform NLI are capable of making---in theory---an infinite number of inferences for a given premise, thereby enabling the testing of a number of pre-defined phenomena of interest. 
For instance, Wang et al. \cite{wang2018glue} formulate many natural language understanding phenomena such as lexical entailment (\textit{dog} entails \textit{animal}), monotonic reasoning (\textit{I do not have a blue book} does not entail \textit{I have no books}), symmetry (\textit{John married Josh} entails \textit{Josh married John} but \textit{Larry likes Sally} does not entail \textit{Sally likes Larry}), etc. as NLI tasks in order to test a range of models.
A caveat in investigating semantic phenomena by studying model predictions on such a task is that the logical relation between the premise and the hypothesis is pre-defined, making it easy for the researcher to relate the model's predictions with its knowledge about the specific phenomena of interest.
However, when it comes to the study of vagueness or fuzziness---which is arguably an integral component of understanding the world, (see \cref{sec:vagueness})---knowing exactly that a temperature of 50$^\circ$ F entails \textit{cool} or \textit{warm} is un-achievable due to the epistemic uncertainty surrounding the concept of vagueness itself \cite{williamson1994vagueness}.
However, the exploration of vagueness within the models' behavioral responses in an NLI setting can still be possible if one supplies multiple plausible hypotheses for a given premise, by relying on commonsense knowledge and intuition.
Comparing the behavior of the model in its predictions on the different hypotheses, then, facilitates roughly exploring the model's encoding of the fuzziness of the given concept.
Revisiting the earlier example, one can compare the extent to which NLI models prefer attributing an entailment relation to a temperature of 50$^\circ$ F with a number of fuzzy categories that may encompass the concept of \textsc{temperature}, such as \textit{warm, cool, moderate,} etc.
In our demonstrative experiments, described below, we test an existing NLI model to classify a natural language premise describing the temperature into five arbitrarily chosen fuzzy categories: \textit{freezing, cold, cool, warm, hot}.

\subsection{Methodology}
\subsubsection{Model investigated}
We perform our experiments on RoBERTa$_{\textit{large}}$ \cite{liu2019roberta}, fine-tuned on the MNLI dataset \cite{williams2018broad}.
The RoBERTa model's core architecture is similar to that of BERT \cite{devlin-etal-2019-bert}, i.e., it uses a transformer trained to bidirectionally predict missing words in context.
Unlike BERT, it does not have the next sentence prediction task, and is trained on 10-times the amount of text data (160 GB) as compared to BERT (16 GB).
To fine-tune on MNLI, the model accepts two sentences, a premise and a hypothesis, in the following format:
\begin{quotation}
\centering
\texttt{\textless cls\textgreater{} Premise \textless/s\textgreater\textless/s\textgreater{}  Hypothesis \textless/s\textgreater},
\end{quotation}
where \texttt{\textless cls\textgreater} is a classification token, and \texttt{\textless/s\textgreater} acts as a sentence separator. 
The model estimates scores (in the form of log-probabilities) for the three classes: entailment, contradiction, and neutral, by applying a linear layer that accepts the representation for the \texttt{\textless cls\textgreater} token and produces a three-dimensional vector with scores for each label. 
We follow this format of representing the input to the model in our experiments.

\renewcommand\arraystretch{1.2}
\setlength\tabcolsep{2.5 pt}
\begin{table}[h!]
\centering
\caption{Configurations used to generate the analytical dataset. \textbf{Note:} The columns ``Location Phrase'' and ``Fuzzy Category'' encompass all rows.}
\vspace{1em}
\label{tab:parameters}
\begin{tabular}{@{}lcccc@{}}
\toprule
\textbf{Unit} & \textbf{\begin{tabular}[c]{@{}l@{}}Temp.\\ Range\end{tabular}} & \textbf{Location Phrase} & \textbf{\begin{tabular}[c]{@{}l@{}}Fuzzy\\ Category\end{tabular}} & \textbf{Total Instances} \\ \midrule
No unit & $[-50, 122]$ & \multirow{3}{*}{\textit{\begin{tabular}[c]{@{}l@{}}No-Location,  in the bedroom,\\ in the living room, in the basement,\\ outside, inside.\end{tabular}}} & \multirow{3}{*}{\textit{\begin{tabular}[c]{@{}l@{}}freezing,\\ cold, cool,\\ warm, hot\end{tabular}}} & 5190 \\
Fahrenheit ($^\circ$F) & $[-50, 122]$ &  &  & 5190 \\
Celsius ($^\circ$C) & $[-50, 50]$ &  &  & 3030 \\ \bottomrule
\end{tabular}
\vspace{-2em}
\end{table}

\subsubsection{Stimuli}
Like any standard NLI input, our stimuli are composed of pairs of premise and hypothesis sentences.
The premise sentences are natural language descriptions of temperature measurements, with the option of a specified location as well as the unit of measurement.
For example, \textit{``It is $n$ degrees outside.''}
For every premise, we construct a hypothesis about each of the plausible fuzzy category to be inferred from the temperature value specified in the premise --- \textit{freezing, cold, cool, warm, hot}. 
For example, \textit{``It is warm in the basement.''}
Our premise and hypothesis templates are shown below, with text in `\{\}' indicating an optional element:
\begin{quotation}
\centering
\textbf{Premise Template:} It is \texttt{[temperature]} degrees \{\texttt{units}\} \texttt{[location-phrase]}.\\
\textbf{Hypothesis Template:} It is \texttt{[fuzzy-category]} \texttt{[location-phrase]}.
\end{quotation}

In summary, we vary between three different unit options, six different location phrases, and five different fuzzy sets in generating our premise-hypothesis pairs, amounting to 54 different settings for every degree of measurement.
The configurations we apply to generate our analytical dataset is presented in \cref{tab:parameters}.  The location phrases are selected opportunistically.
In total we test on 13{,}410 different sentence pairs.


\subsubsection{Analysis and Results}
To examine the model's behavior in making fuzzy inferences from a given natural language description of a temperature setting, we compute the probability the model assigns to the ``entailment'' label for every premise-hypothesis pair.
Formally, for a given pair $\langle P_i, H_{ij}\rangle$, where $i$ is the temperature in the premise, and $j$ is the perceived fuzzy category specified in the hypothesis, we compute the entailment score $\ee$ as:
\begin{align}
    \ee = p(\text{``entailment''} \mid \langle P_i, H_{ij}\rangle).
\end{align}
Since this computation is probabilistic, we cannot translate it directly into a membership function.
We instead focus on the patterns the model shows in its entailment scores for all the fuzzy categories we consider, in order to characterize its behavior in terms of how well it can mimic qualitative properties of membership functions, both typical and atypical.
\Cref{tab:qualitative} shows two randomly selected premises, their corresponding hypotheses, and $\ee$-values computed using the model that we study.

\begin{figure}[p!]
\centering
\subfloat[\label{fig:nounit}]{%
  \includegraphics[clip,width=0.76\textwidth]{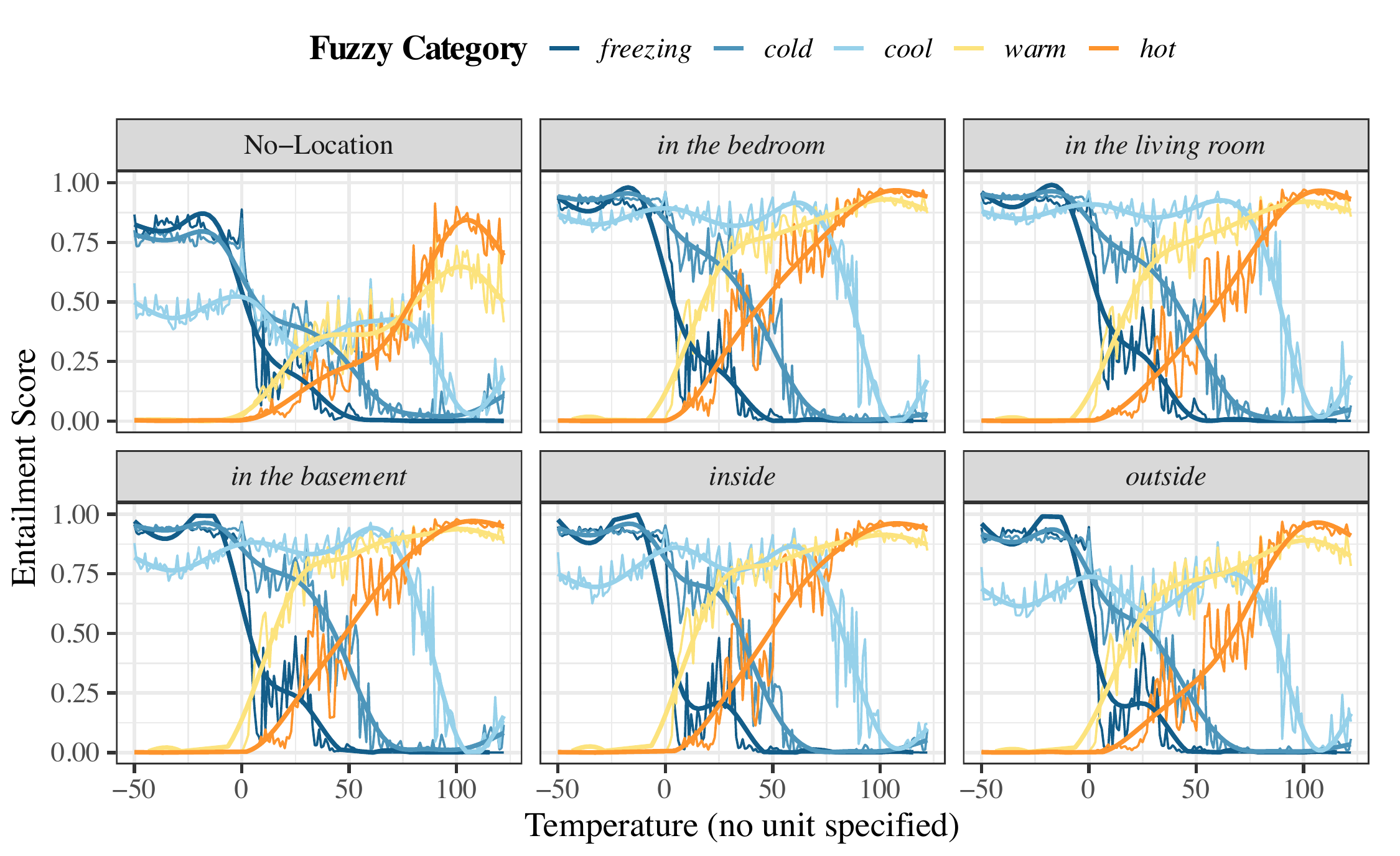}%
}

\subfloat[\label{fig:fahrenheit}]{%
  \includegraphics[clip,width=0.76\textwidth]{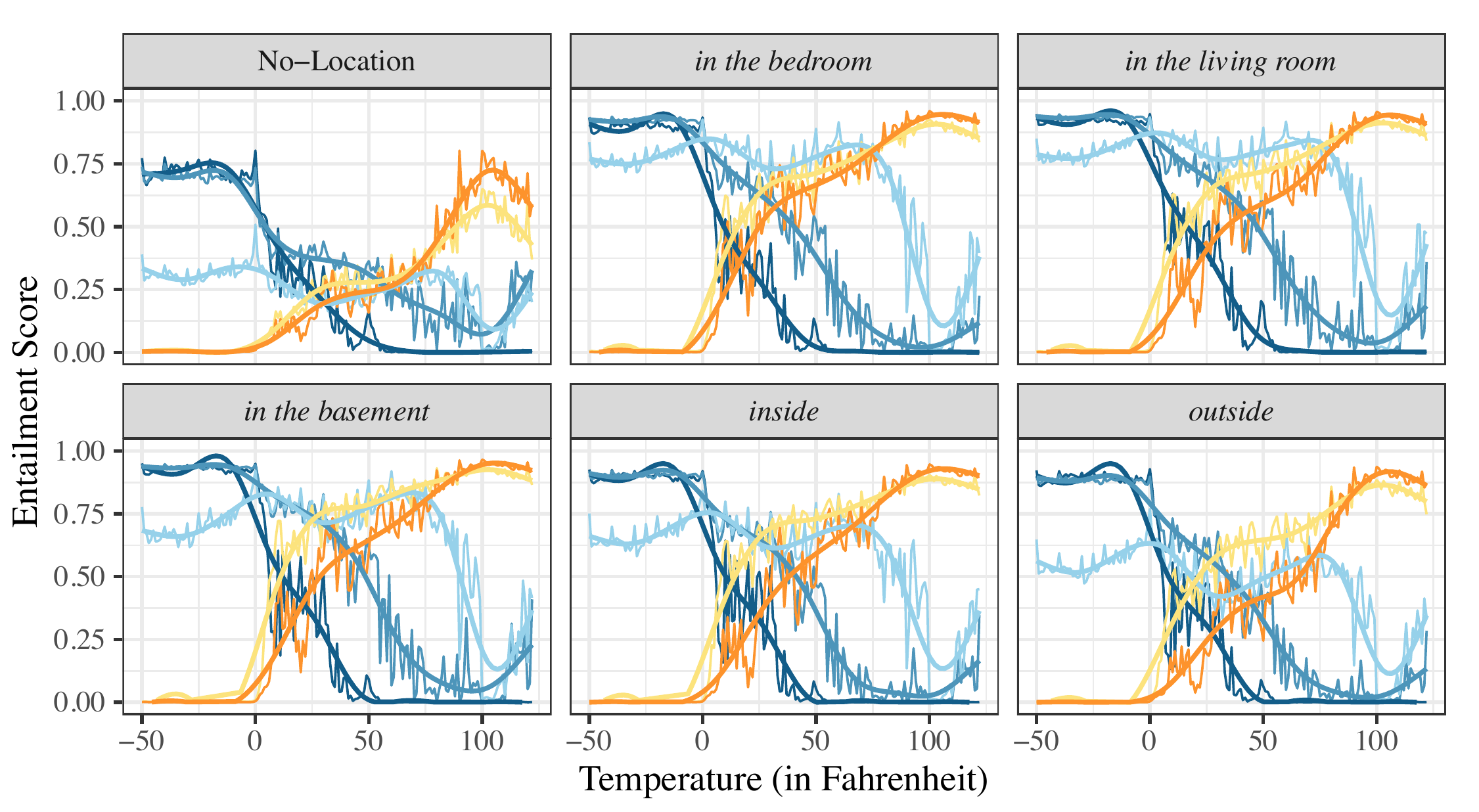}%
}

\subfloat[\label{fig:celsius}]{%
  \includegraphics[clip,width=0.76\textwidth]{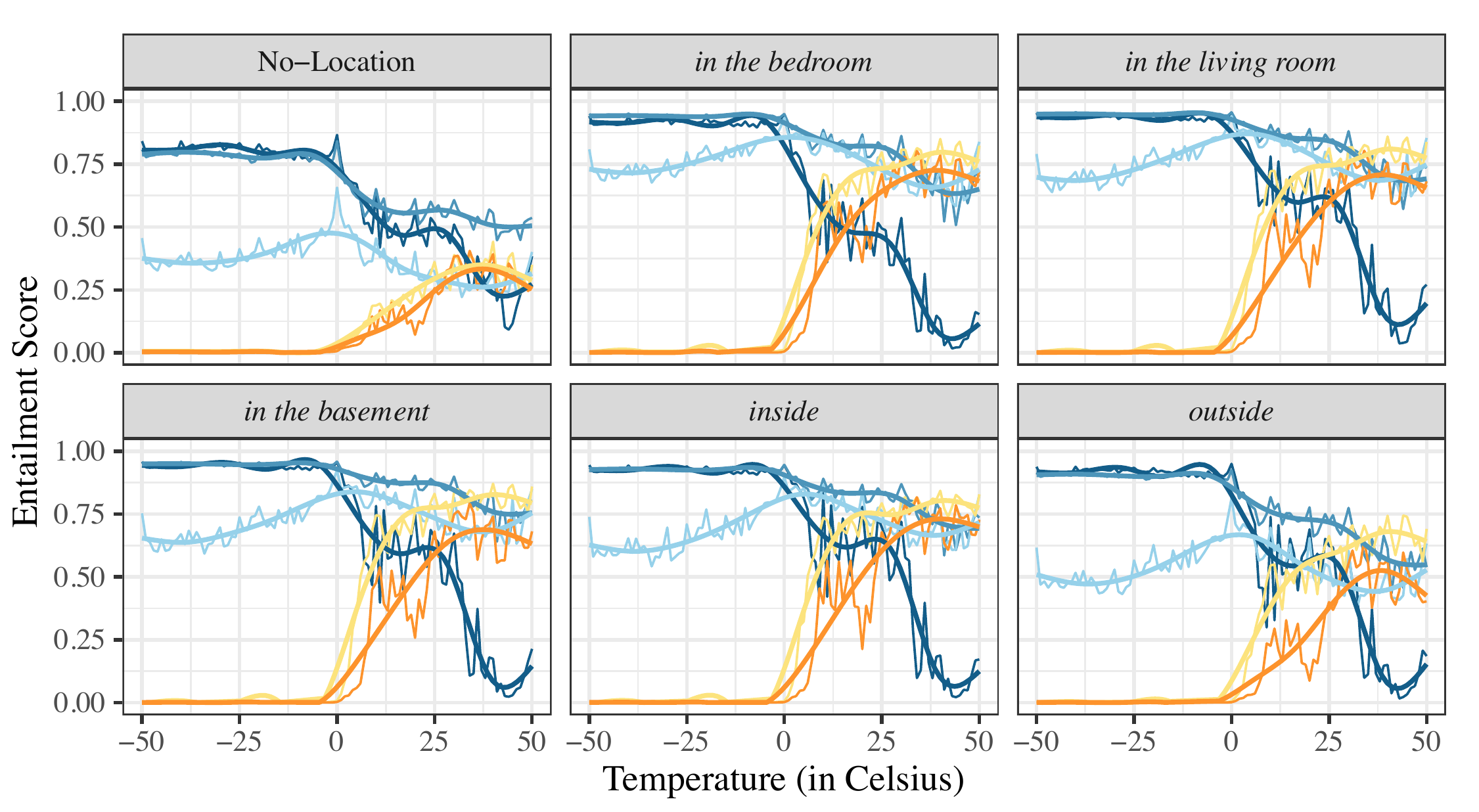}%
}

\caption{Entailment scores (y-axis) from RoBERTa$_\textit{large}$ for various fuzzy categories, given premise with specific temperature (x-axis): \textbf{(a)} No unit; \textbf{(b)} Fahrenheit; and \textbf{(c)} Celsius. Scores are smoothed using a generalized additive model (with default parameters, in R).}
\label{fig:temperature}
\end{figure}

The results of our experiments are collectively shown in \cref{fig:temperature}.
Each plot in \cref{fig:temperature} represents the model's $\ee$-values for each of the five hypotheses types in a given unit of measurement and location setting.
To better understand general patterns in each setting, we smooth out the noise in our measurements by fitting a Generalized Additive Model (GAM) \cite{hastie1990generalized}.
The first thing to note from our results is that, overall, model shows noisy but qualitatively sound numeric knowledge patterns based on how the raw temperature values relate to the model's synthetic perception --- large values typically show greater $\ee$-values for the \textit{warm} and \textit{hot} hypotheses than for hypotheses representing \textit{freezing, cold,} and \textit{cool}.
This is especially noticeable upon comparing no-unit (\cref{fig:nounit}) and Fahrenheit (\cref{fig:fahrenheit}) conditions with that of Celsius (\cref{fig:celsius}), we see that the model has approximately grasped what is considered to be ``cold'' in the Celsius condition --- the model shows almost zero-membership values for \textit{warm} and \textit{hot} when the temperature specified in the premise is below 0$^\circ$C and gradually goes up to positive $\ee$-values as the temperature increases.
In the Fahrenheit condition, however, the model is unable to show exactly the same pattern (0$^\circ$C $=$ 32$^\circ$F), suggesting a slight lack of internal consistency.
\renewcommand\arraystretch{1.2}
\setlength\tabcolsep{2.5 pt}
\begin{table}[t!]
\centering
\caption{Examples of the extent to which RoBERTa$_{\textit{large}}$---fine-tuned on the MNLI dataset \cite{williams2018broad}---is able to infer one of five arbitrarily chosen temperature categories when given the temperature of a given room (no unit specified). \textbf{Note:} ``Entailment Scores'' are not the same as fuzzy membership values, and are instead used to perform the behavioral analysis of the model.}
\vspace{1em}
\label{tab:qualitative}
\begin{tabular}{@{}llc@{}}
\toprule
\textbf{Premise} & \textbf{Hypotheses} & \textbf{Entailment Score} ($\ee$) \\ \midrule
\multirow{5}{*}{\textit{It is 0 degrees in the bedroom.}} & \textit{It is \textbf{freezing} in the bedroom.} & 0.956 \\
 & \textit{It is \textbf{cold} in the bedroom.} & 0.961 \\
 & \textit{It is \textbf{cool} in the bedroom.} & 0.962 \\
 & \textit{It is \textbf{warm} in the bedroom.} & 0.009 \\
 & \textit{It is \textbf{hot} in the bedroom.} & 0.004 \\ \midrule
\multirow{5}{*}{\textit{It is 70 degrees in the living room.}} & \textit{It is \textbf{freezing} in the living room.} & \textless 0.001 \\
 & \textit{It is \textbf{cold} in the living room.} & 0.002 \\
 & \textit{It is \textbf{cool} in the living room.} & 0.928 \\
 & \textit{It is \textbf{warm} in the living room.} & 0.902 \\
 & \textit{It is \textbf{hot} in the living room.} & 0.713 \\ \bottomrule
\end{tabular}
\end{table}
Revisiting the Celsius condition, the model is rather noisy, as compared to its no-unit and Fahrenheit counterparts --- it shows similar $\ee$-values for \textit{cool, cold, warm,} and \textit{hot} categories in this condition, deviating slightly from intuitive perceptions about temperature. 
For instance, at 50$^\circ$C, the perception of the model leans more towards \textit{cold} than \textit{hot}.
The model's patterns when no unit of measurement is specified in the premise show strong correspondence with that in the Fahrenheit condition (RMSE\footnote{root mean-squared error} between No-unit and Fahrenheit is 0{.}09, while that between No-unit and Celsius is 0{.}16).
This suggests that when no information is available, the model defaults to inferring Fahrenheit as the unit of measuring temperature.
We hypothesize that this is likely because the model is primarily trained on American-influenced text (e.g. Reddit).

\begin{figure}
    \centering
    \includegraphics[width=\textwidth]{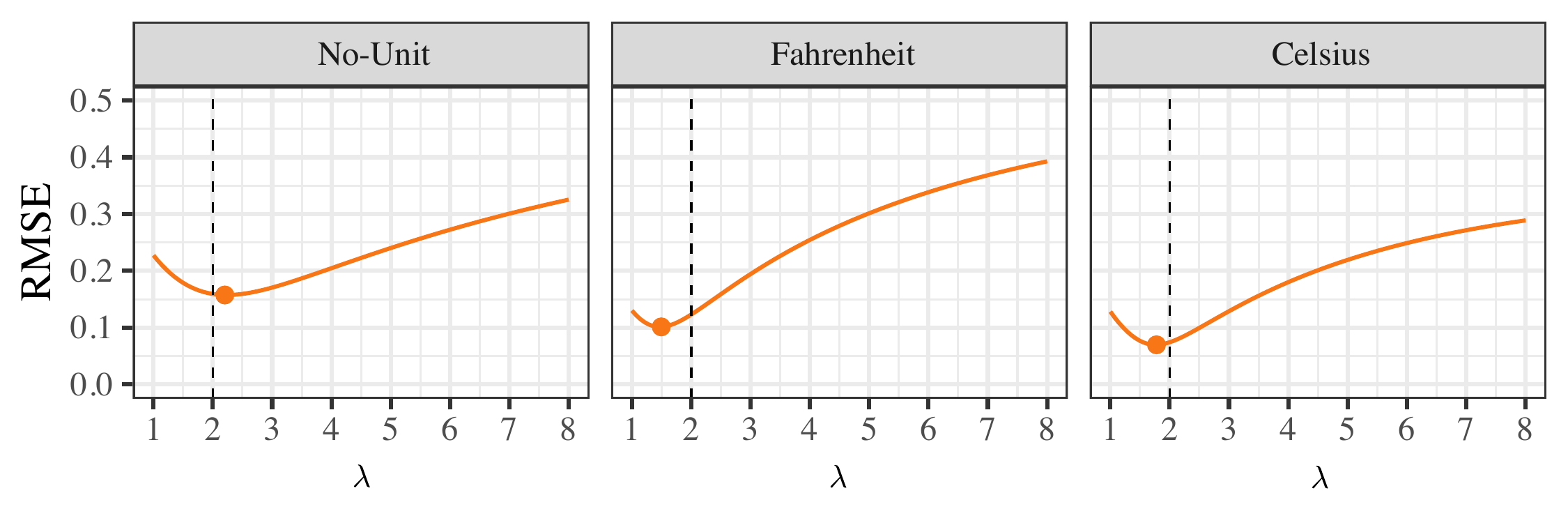}
    \caption{Points representing the best $\lambda$ values for minimizing $\textsc{rmse}(\ee_{\textit{hot}}, \ee^\lambda_{\textit{warm}})$ using a na\"ive search among 100 values in $[1, 8]$. Dashed line represents $\lambda$ corresponding to \textit{very} as per Zadeh \cite{zadeh1972fuzzy}.}
    \label{fig:rmse}
\end{figure}

In the context of computational treatment of fuzziness, we find the model to show strikingly noticeable similarities to fuzzy hedges \cite{lakoff1975hedges, zadeh1972fuzzy}.
More prominently, the $\ee$-values for \textit{hot} bear qualitative resemblance to a `concentration' transformation of the $\ee$-values for \textit{warm}.
That is, if we take the membership of \textit{warm} and raise it to the power of some value greater than 1 (say, $\lambda$), we should get a hedged version of \textit{warm} ($H =$ \textit{very, extremely}, etc.)  that could potentially be mapped to the \textit{hot} category, pulling the curve inward, as proposed by Zadeh \cite{zadeh1972fuzzy}:
\begin{align}
    \textit{hot} \approx H \textit{ warm} = \int_U \mu^\lambda(x) \tag{$\lambda > 1$}
\end{align}
Empirically, we perform a na\"ive search between $[1, 8]$ (100 values) and minimize the RMSE between $\ee_{\textit{hot}}$ and $\ee^\lambda_{\text{warm}}$ to find the best value for $\lambda$ as per the above equation, and show results in \cref{fig:rmse}.
We find the best $\lambda$ value to be close to 2 in all three cases (with Fahrenheit showing the greatest deviation), suggesting that \textit{hot} is approximately equal to \textit{very warm} (if Zadeh's recommendation in \cite{zadeh1972fuzzy} is to be followed for \textit{very}).
We also find the $\ee$-value for \textit{warm} to be greater than that of \textit{hot} approximately 77.64\%, 75.33\%, and 98.51\% of the time for the no-unit, Fahrenheit, and Celsius conditions, respectively.
This empirically supports the above notion that highlights the possibility of hedging \textit{warm} to produce \textit{hot}, and also largely follows the `fuzzy' interpretation of semantic entailment ($P$ (\textit{hot} or hedged-\textit{warm}) entails $Q$ (\textit{warm}) iff $\mu_P(x) \leq \mu_Q(x)$), as noted by \cite{lakoff1975hedges}, with its strongest effect being included in the Celsius condition.
Finally, we also observe a tendency of the \textit{hot} and \textit{warm} curves to show a decrease in extremely hot temperatures, suggesting the existence of another fuzzy set that is applicable to more extreme temperatures.
This is likely to be an antonym of \textit{freezing} or something that is synonymous to \textit{very hot}.
However, since there isn't a clear word with comparable frequency,\footnote{we experimented with \textit{scorching}, \textit{blistering}, and \textit{balmy}, but all three have considerably greater differences in frequency with \textit{freezing}, as compared to that between \textit{cold}-\textit{hot} and \textit{cool}-\textit{warm}. This can be verified by using a public resource such as Google's n-gram viewer.} we leave this analysis for future work.

\section{General Discussion and Conclusion}
More often than not, natural language expressions have a graded---rather than absolute---measurement of truth and falsehood \cite{lakoff1975hedges}.
This graded property of language owes itself to the vague boundaries in concepts such as \textsc{tall}, \textsc{fruit}, etc., that do not have a clear-cut criteria for membership.
The development of fuzzy set theory \cite{zadeh1965fuzzy} has provided us with a useful toolkit to handle imprecision in language, e.g. ``\textit{the coffee is \underline{very warm}},'' by formulating membership functions to model data.
We argue in this paper that computational models---such as PLMs---that learn language representations must account for the fuzziness contained in language use.
To this end, we formulated a demonstrative experiment by investigating a current state of the art model (RoBERTa$_\textit{large}$) for the classical fuzzy case of \textsc{temperature}---to what extent can it map natural language premises such as \textit{``it is 40$^\circ$F inside''} to hypotheses carrying fuzzy categories such as \textit{``it is cold inside''}
Our tests follow the task of natural language inference \cite{bowman2015large}, which allows broad-coverage testing of a variety of semantic phenomena encoded in models.
Furthermore, the model we test was fine-tuned to perform the NLI task, making our investigation a natural environment for the model.
We primarily focus on our model's behavioral responses to the supplied premise-hypotheses pairs by examining its entailment-scores (the degree to which the premise entails the hypothesis, according to the model) for various fuzzy categories mentioned in the hypotheses, and analyse their patterns.
Overall, we found the model to show a noisy-but-sound grasp of temperature perception, with substantial differences between cold and warm temperature categories towards the extreme ends of the scale.
While the model showed sound patterns of temperature perception across our unit-conditions, we found it to show slightly lower correspondence with intuition when the temperature mentioned in the premise was in Celsius.
This, together with the fact that its patterns on the Fahrenheit condition are found to be strongly similar to that in the no-unit condition suggest that the model has a  bias towards American English.
In our main analyses, the model showed similar patterns in its entailment scores across the fuzzy categories as standard membership functions formulated for showing hedging effects. 
For instance, the model's entailment scores for the category of \textit{hot} mimicked the entailment-scores of a concentrated version of \textit{warm}, pulling its curve inward. However this was not the case in the colder temperatures, suggesting a lack of consistency in this behavior.
This indicates, tentatively, that state of the art models show noticeable signs of approximately showing patterns akin to that of fuzzy membership functions in graded cases such as \textsc{temperature}.
We therefore use this paper as a call towards a systematic study of the manifestation of vagueness in state of the art language models, as well as how multi-valued logic such as fuzzy logic can be incorporated within them in order to better handle imprecise language.

\vspace{1em}
\noindent
{\colorbox{yellow}{\textbf{Disclaimer:}}} This is a preprint of the accepted manuscript: Kanishka Misra and Julia Taylor Rayz, Finding fuzziness in Neural Network Models of Language Processing, to be presented at NAFIPS 2021, whose proceedings will be published in \textit{Explainable AI and Other Applications of Fuzzy Techniques}, edited by Julia Taylor Rayz, Victor Raskin, Scott Dick, and Vladik Kreinovich, reproduced with permission of Springer Nature Switzerland
AG. The final authenticated version is available online at: \texttt{url TBD}.

\vspace{1em}
\noindent
\textbf{Reproducibility:} The code to reproduce experiments reported in this paper can be found at \texttt{\url{https://github.com/kanishkamisra/nafips2021}}.



%
%
\bibliographystyle{spmpsci}
\bibliography{references.bib}

\end{document}